\documentclass[11pt]{article}

\usepackage[final]{acl}

\usepackage{times}
\usepackage{latexsym}

\usepackage[T1]{fontenc}

\usepackage[utf8]{inputenc}

\usepackage{microtype}

\usepackage{inconsolata}

\usepackage{graphicx}

\usepackage{booktabs}

\usepackage{amsmath} 
\usepackage{amsfonts}
\usepackage{multirow}

\title{STRETCH the Boundaries: A Unified Self-Taught Framework for Progressive LLM Evolution}

\author{Yajie Yu \\
School of Computer Science\\ University of Birmingham\\
Birmingham\\ United Kingdom \\
\texttt{yxy616@student.bham.ac.uk}\\\And
Mark Lee\\
School of Computer Science\\ University of Birmingham\\
Birmingham\\ United Kingdom \\
\texttt{m.g.lee@bham.ac.uk}\\\And
Yue Feng\thanks{Corresponding author.}\\
School of Computer Science\\ University of Birmingham\\
Birmingham\\ United Kingdom \\
\texttt{y.feng.6@bham.ac.uk}
}

\begin{document}
\maketitle
\begin{abstract}
Large language models (LLMs) often suffer from capability stagnation in self-improvement training because fixed difficulty levels fail to adapt to their evolving proficiency. To address this issue, we propose \textbf{STRETCH}  (\textbf{S}elf-\textbf{T}aught \textbf{R}easoning \textbf{E}volution via \textbf{T}argeted \textbf{CH}allenge), a unified framework inspired by cognitive scaffolding theory. STRETCH introduces a dynamic Stretch Zone mechanism that continuously aligns question difficulty with the model’s solving capability. Within a single parameter space, the model alternates between a Scaffolder that generates adaptive, boundary-pushing challenges and a Learner that that optimizes its solving trajectories through reinforcement learning. This dual-loop co-evolution effectively stabilizes training, mitigates reward hacking and promote progressive reasoning growth. Experiments on both negotiation and operation research benchmarks demonstrate that STRETCH consistently outperforms strong prompting and domain-specific baselines. Further scaffolder configuration analysis shows that dynamic difficulty alignment is critical for sustained capability improvement and synchronized reasoning evolution.\footnote{Our code is available at \url{https://github.com/GuanNiPiShi123/STRETCH}.}
\end{abstract}
\section{Introduction}
\label{sec:intro}
\begin{figure}[!t]
    \centering
    \includegraphics[width=\columnwidth]{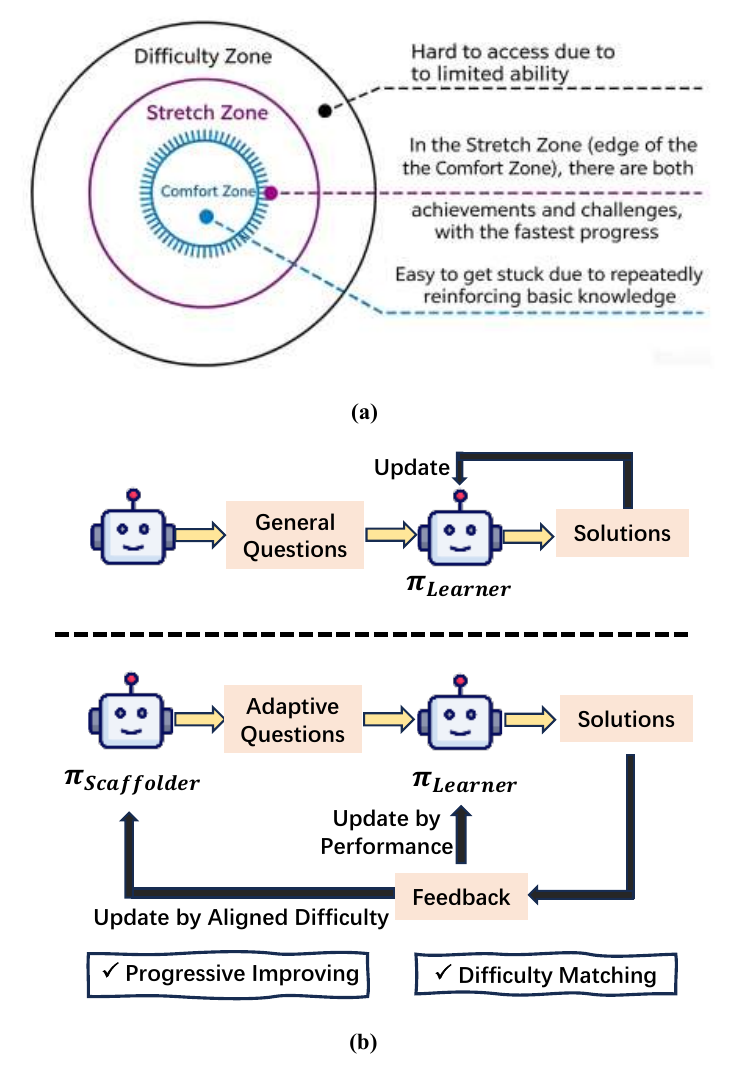}
\caption{
        Motivation of the STRETCH Framework.
        \textbf{(a) Cognitive Zones:} Comfort, stretch (optimal learning), and difficulty.
        \textbf{(b) Paradigm Comparison:}
        Top: static general questions leading to learner capability stagnation. 
        Bottom: adaptive questions achieving dynamic difficulty matching via a learnable Scaffolder for progressive improving of Learner.
    }
    \label{fig:stretch_motivation}
\end{figure}
Large language models (LLMs) have demonstrated remarkable capabilities in complex reasoning domains, ranging from mathematical problem-solving to strategic social interactions \citep{sicilia-etal-2024-deal, yu2025policyevolagentevolvingpolicyenvironment, patil2025advancingreasoninglargelanguage,zeng-etal-2025-dynamic, wu-etal-2025-agentic, liu2026agenticpay}. To push the boundaries of these models beyond imitation learning from human-annotated data, recent advancements have increasingly adopted post-training paradigms based on self-play exploration mechanism \citep{zhao2025absolutezeroreinforcedselfplay, guo2025genenvdifficultyalignedcoevolutionllm,zhao2026insideoutevolvingusercentric,liu2026position}. Despite these promising strides, current self-improvement frameworks face a critical bottleneck: they largely lack a mechanism for progressive capability improvement. Traditional reinforcement learning (RL) approaches typically benchmark models on static datasets or predefined difficulty levels \citep{schulman2017proximalpolicyoptimizationalgorithms,rafailov2024directpreferenceoptimizationlanguage,lee2025evolvingdeeperllmthinking, wei2026evomemorybenchmarkingllmagent}. Under such static setups, models easily get ``stuck'': as the LLM masters the fixed data distribution, it quickly falls into a capability plateau, leading to premature convergence. Conversely, if initial tasks are excessively complex, the agent fails to find valid reasoning trajectories, resulting in catastrophic reward collapse.

To systematically overcome this stagnation, we draw inspiration from cognitive scaffolding theory \citep{wilson2014scaffolding,article_scaffolding, newen2025pattern}, which partitions human learning into three cognitive territories (as illustrated in Figure~\ref{fig:stretch_motivation}a). For optimal learning, an agent should avoid the Comfort Zone (where repeatedly reinforcing basic knowledge leads to stagnation) and the Difficulty Zone (where tasks are inaccessible due to limited ability). Instead, learning must be anchored in the \textbf{Stretch Zone}---a dynamic cognitive boundary just beyond the Comfort Zone where achievements and challenges fluidly coexist. As shown in the paradigm comparison in Figure~\ref{fig:stretch_motivation}b (Top), traditional training paradigms fail because they rely on static general questions; as the model grows stronger, the Stretch Zone shifts upward, but the environment remains fixed, dropping the model back into the Comfort Zone. Breaking through early plateaus necessitates a dynamic paradigm that performs continuous difficulty matching (Figure~\ref{fig:stretch_motivation}b, Bottom), ensuring the generation of adaptive questions that scale in tandem with the model's growing proficiency.

To achieve this cognitive-inspired alignment, we propose \textbf{STRETCH}, a unified self-taught framework for progressive LLM evolution. Unlike prior curriculum learning or self-play methods that rely on frozen environment generators or multiple disjoint models \citep{yang2026ttcs,karlekar2026duelevolverewardfreetesttimescaling}, STRETCH coordinates a unified large language model to seamlessly alternate between two complementary cognitive roles within a singular parameter space. Acting as a learnable \textbf{Scaffolder}, the model utilizes historical outcome feedback to dynamically generate adaptive multi-task constraints (e.g., strategic negotiation bottom lines or complex mathematical resource limits) that reside strictly within the model's current Stretch Zone. Subsequently, switching to the \textbf{Learner} role, the model interacts with external environments to navigate these self-generated challenges. Because both roles share the same underlying weights, optimizing the Learner's trajectories via a hybrid pipeline of Group Relative Policy Optimization (GRPO) \citep{shao2024deepseekmathpushinglimitsmathematical,zhang2026critiquegrpoadvancingllmreasoning} enhances the Scaffolder's strategic capacity to propose realistic, perfectly aligned challenges. This joint parametric co-evolution creates a continuous self-improvement loop.

We empirically evaluate the cross-domain generalization capability of STRETCH on two highly divergent constrained reasoning benchmarks: multi-turn bargaining (the Craigslist dataset) and advanced mathematical operation research (the Mano Complex and Complex OR datasets). Experimental results demonstrate that STRETCH consistently breaks through the early capability plateaus that shackle traditional static RL, significantly outperforming state-of-the-art domain-specific baselines across both social game play and rigid mathematical optimization. 

Our main contributions are as follows:
\begin{itemize}
    \item \textbf{Cognitive-Inspired Diffculty Alignment:} We formalize the post-training of LLMs through cognitive scaffolding theory. By conceptualizing the Stretch Zone, we establish a concrete framework that utilizes continuous difficulty matching to overcome capability stagnation of model.
    \item \textbf{Unified Parametric Co-evolution:} We propose STRETCH, characterized by the simultaneous evolution of both the task-generating Scaffolder and the task-solving Learner within a single parameter space, eliminating the need for disjoint models.
    \item \textbf{Cross-Domain Breakthrough:} We design a robust self-taught loop powered by feedback-driven task generation, achieving significant performance gains on both strategic negotiation and operation research tasks, proving the efficacy of our evolution framework.
\end{itemize}
\section{Related Works}
\begin{figure*}[!t]
    \centering
    \includegraphics[width=\linewidth]{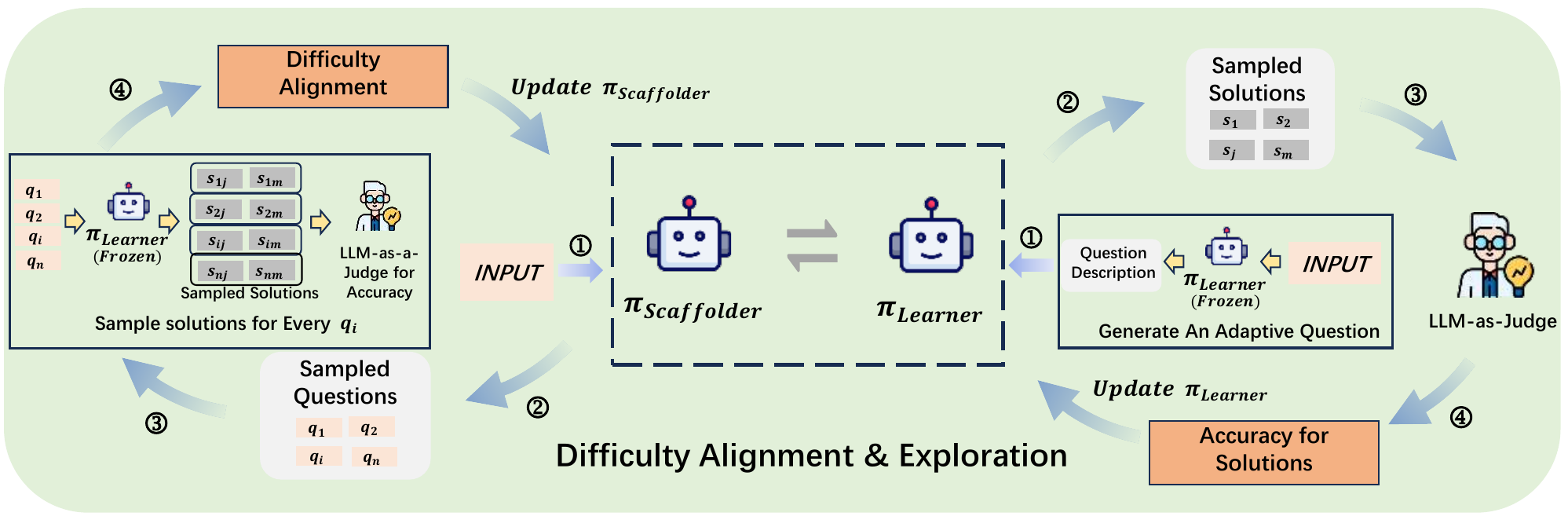}
\caption{The dual-loop asynchronous updating mechanism of STRETCH. 
\textbf{Difficulty Alignment (Left Loop)}:$\pi_{Scaffolder}$ is updated to generate questions within the Learner's Stretch Zone by driving a frozen Learner to sample solutions and evaluating its empirical accuracy. 
\textbf{Exploration (Right Loop):} $\pi_{Learner}$ is updated to solve adaptive questions generated by a frozen Scaffolder.}
    \label{fig:method}
\end{figure*}
\subsection{Co-Evolutionary Frameworks}

Self-play has emerged as a promising paradigm for LLM self-improvement, where models generate and solve their own problems. \citet{zhao2025absolutezeroreinforcedselfplay} proposed Absolute Zero Reasoner (AZR), a self-play reasoning framework that operates without any external human-annotated or distillation data. In parallel, \citet{shao2024deepseekmathpushinglimitsmathematical} introduced DeepSeekMath, which leverages self-generated data for mathematical reasoning and introduces Group Relative Policy Optimization (GRPO) to enhance reasoning abilities while optimizing memory usage.

Multiple studies have explored the joint evolution of task generators and solvers \citep{chen2026spc,xiang2026systematic,luo2026storageexperiencesurveyevolution}.\citet{guo2025genenvdifficultyalignedcoevolutionllm} proposed GenEnv, a difficulty-aligned co-evolution framework between an LLM agent and a scalable environment simulator. \citet{huang2025rzero} introduced R-Zero, which separately optimizes a Challenger and a Solver initialized from a common base model. \citet{yang2026ttcs} introduced TTCS, a test-time curriculum synthesis framework where a question synthesizer and a reasoning solver co-evolve from the same pretrained model. \citet{sygkounas2026covolveadversarialcoevolutionlargelanguagemodelgenerated} proposed COvolve, which models the interaction between environment and policy designers as a two-player zero-sum game, ensuring adversarial co-evolution. \citet{huang2026gzeroselfplayopenendedgeneration} presented G-Zero, a verifier-free co-evolutionary framework that drives continuous self-evolution through hint-induced response.

\subsection{Cognitive Scaffolding in LLMs}
Drawing from cognitive science, the concept of scaffolding has been increasingly applied to LLM reasoning. Cognitive scaffolding theory originates from the concept of the Zone of Proximal Development (ZPD) \citep{article_sc}. \citet{wilson2014scaffolding} provided a foundational articulation of scaffolding as ``high challenge, high support'', arguing that effective scaffolding enables learners to achieve far beyond what they could accomplish individually. \citet{newen2025pattern} developed a pattern theory of scaffolding, characterizing how environmental resources contribute to the realization of mental abilities and offering a framework for understanding the functional role of scaffolding in cognitive systems. \citet{article_scaffolding} showed how scaffolding regulates the flow of information within the learner's working memory, thereby reducing cognitive load.

Recent work has operationalized this theory for LLMs \citep{chen2026expanding,cui-sachan-2025-investigating,wallis2026llmszpd}. Inspired by cognitive scaffolding, \citet{kim2025guidingreasoningsmalllanguage} introduced SMART (Small Reasons, Large Hints), where LLMs provide targeted, selective guidance to augment small language model reasoning. Another line of research explores how prompt-level inductive biases serve as cognitive artifacts. \citet{li2025stayingsweetspotresponsive} proposed SEELE, a supervision-aided RLVR framework that dynamically adjusts problem difficulty by appending hints of adaptive length, keeping rollout accuracy near the theoretically optimal 50\% regime.
\section{Methodology}

To overcome the capability stagnation prevalent in static reinforcement learning, we introduce STRETCH, a progressive self-evolution framework. As illustrated in Figure~\ref{fig:method}, STRETCH employs a singular parameter space that seamlessly alternates between two cognitive roles: the Scaffolder ($\pi_{Scaffolder}$) and the Learner ($\pi_{Learner}$). Our dual-loop asynchronous updating mechanism structurally enforces continuous difficulty alignment and progressive exploration.

At a high level, STRETCH optimizes task validity, learnability, and replay-stabilized downstream behavior jointly:
\begin{equation}
\max_{\theta}\; \mathbb{E}_{q\sim\pi_\theta}\!\left[\mathbb{I}_{\mathrm{valid}}(q)R_{\mathrm{diff}}(q)\right] + \lambda \mathcal{R}_{\mathrm{replay}}(\theta),
\label{eq:global_objective}
\end{equation}
where $\mathbb{I}_{\mathrm{valid}}(q)$ is supplied by the domain verifier, $R_{\mathrm{diff}}$ favors tasks at the Learner's capability boundary, and $\mathcal{R}_{\mathrm{replay}}=-\mathcal{L}_{\mathrm{SFT}}$ rewards likelihood on verified successful trajectories. Equation~\ref{eq:global_objective} is implemented by the alternating GRPO updates and epoch-level replay described below, rather than optimized as a single differentiable loss.

\subsection{Role Switching within a Singular Space}
\label{subsec:role_switching}

Unlike traditional co-evolution frameworks that require maintaining and syncing two separate large language models \citep{guo2025genenvdifficultyalignedcoevolutionllm,yang2026ttcs, zhou2026steporlm}, STRETCH unifies task generation and task solving within a single base model $\pi_\theta$. The dual roles are activated purely via role-specific system prompts. 

To enable stable mutual learning, we implement an alternating phase-update mechanism. During the Difficulty Alignment (Left Loop), we optimize the active Scaffolder $\pi_{Scaffolder}$ while keeping a historical snapshot of the Learner frozen. Conversely, during the Exploration (Right Loop), we optimize the active Learner $\pi_{Learner}$ against constraints generated by the frozen Scaffolder. This decoupling loop ensures that each updating iteration receives consistent and reliable reward signals.

\subsection{Scaffolder Optimization}
\label{subsec:left_loop}

The primary objective of the left loop is to train the Scaffolder to dynamically discover the Learner's Stretch Zone. It must learn to propose questions that are neither trivially easy (falling into the Comfort Zone) nor impossibly hard (falling into the Difficulty Zone.

\noindent \textbf{Group Question Generation.} Given a base task context $c$, the active Scaffolder $\pi_{Scaffolder}$ samples a group of $n$ candidate adaptive questions:
\begin{equation}
    Q = \{q_1, q_2, \dots, q_m\} \sim \pi_{Scaffolder}(\cdot \mid c)
\end{equation}

\noindent \textbf{Frozen Learner Rollout.} To empirically measure the difficulty of these generated questions, we utilize the frozen Learner $\pi_{Learner}^{(\text{Frozen})}$. For each candidate question $q_i$, the frozen Learner independently samples $n$ diverse reasoning trajectories (solutions):
\begin{equation}
    S_i = \{s_{i1}, s_{i2}, \dots, s_{im}\} \sim \pi_{Learner}^{(\text{Frozen})}(\cdot \mid q_i)
\end{equation}

\noindent \textbf{Difficulty Alignment Reward.} An external LLM-as-a-Judge evaluates the correctness of each solution $s_{ij}$. The empirical success rate (accuracy) for a given question $q_i$ is computed as $\hat{p}_i = \frac{1}{m} \sum_{j=1}^{m} \mathbb{I}(\text{correct})$. To explicitly force the Scaffolder to target the Stretch Zone, we formulate a difficulty-alignment reward that peaks at 50\% success rate:
\begin{equation}
    R_{diff}(q_i) = \left(\hat{p}_i  \left(1-\hat{p}_i\right)\right)^{\alpha}
\end{equation}

where $\alpha$ curves the reward sharpness. Since a Bernoulli outcome with success probability $p$ has variance $p(1-p)$, this reward is maximized at $p=0.5$: the regime with the most informative success/failure feedback for relative-policy updates. Validity is not inferred from the target rate alone: infeasible OR constraints receive zero verifier reward, while negotiation trajectories are replay-regularized using successful, rule-consistent dialogues.
This reward penalizes the Scaffolder if the frozen Learner effortlessly solves all samples ($\hat{p}_i \to 1$) or fails entirely ($\hat{p}_i \to 0$).

\noindent \textbf{GRPO Update for Scaffolder.} We normalize the rewards within the group of $m$ questions to compute the advantages $A_j$, and then update Scaffolder's parameters $\theta$ using the GRPO objective to maximize the probability of generating difficulty-aligned questions:
\begin{equation}
\begin{aligned}
    \mathcal{L}_{Scaffolder}(\theta) &= \mathbb{E} \Bigg[ \sum_{j=1}^m \min \Big(  r_j(\theta) A_j, \\
    & \text{clip}\big(r_j(\theta), 1-\epsilon, 1+\epsilon\big) A_j \Big) \Bigg]
\end{aligned}
\end{equation}
where $r_j(\theta) = \frac{\pi_{Scaffolder}(q_j \mid c)}{\pi_{old}(q_j \mid c)}$ is the probability ratio for the Scaffolder’s responses.

\subsection{Learner Optimization}
\label{subsec:right_loop}

After the Scaffolder's parameters are updated to target the model's cognitive boundaries, we freeze it and shift to the right loop. The objective here is to advance the Learner's reasoning capabilities to conquer newly generated constraints.

\noindent \textbf{Adaptive Constraint Specification.} The newly frozen Scaffolder $\pi_{Scaffolder}^{(\text{Frozen})}$ is prompted to generate a specific, difficulty-aligned question $q_{adapt}$ based on the training context.

\noindent \textbf{Group Solution Sampling.} The active Learner $\pi_{Learner}$ interacts with $q_{adapt}$ to sample a group of $m$ candidate reasoning trajectories:
\begin{equation}
    S = \{s_1, s_2, \dots, s_m\} \sim \pi_{Learner}(\cdot \mid q_{adapt})
\end{equation}

\noindent \textbf{Accuracy Evaluation and GRPO Update.} The LLM-as-a-Judge evaluates each trajectory $s_j$ for logical correctness and constraint adherence, assigning an accuracy-based reward $R_{acc}(s_j)$. For OR tasks, $R_{acc}$ is a binary reward strictly based on code execution and optimality. For Negotiation tasks, $R_{acc}$ is a joint score assigned by the LLM-as-a-Judge evaluating deal success and strategic consistency. We compute the relative advantages $A_j$ within this group of $m$ solutions. The Learner's parameters $\theta$ are then updated via the GRPO objective to master the current Stretch Zone:
\begin{equation}
\begin{aligned}
    \mathcal{L}_{\text{Learner}}(\theta) &= \mathbb{E} \Bigg[ \sum_{j=1}^m \min \Big(  \rho_j(\theta) A_j, \\
    & \text{clip}\big(\rho_j(\theta), 1-\epsilon, 1+\epsilon\big) A_j \Big) \Bigg]
\end{aligned}
\end{equation}
where $\rho_j(\theta) = \frac{\pi_{Learner}(s_j \mid q_{adapt})}{\pi_{old}(s_j \mid q_{adapt})}$ is the probability ratio for the Learner's trajectories.

\subsection{The Progressive Spiral with Golden Experience Replay}
\label{subsec:spiral}

To prevent either cognitive role from over-optimizing or collapsing, the dual objectives of STRETCH are integrated into a micro-interleaved optimization loop. At each training step, the framework performs exactly one Scaffolder update followed immediately by one Learner update, culminating in an epoch-level Golden Experience Replay to solidify behavioral patterns.

Concretely, the execution flow is structured as:

\textbf{Micro-Interleaved GRPO Updates for Scaffolder and Learner (Step-Level).}
\begin{itemize}
        \item \textit{Scaffolder Turn:} The active Scaffolder $\pi_{Scaffolder}$ samples questions, where the shared parameters $\theta$ are updated to push the difficulty boundary.
        \item \textit{Learner Turn:} The active Learner $\pi_{Learner}$ samples solutions against the updated constraints, and $\theta$ is updated again via GRPO to master the new difficulty.
    \end{itemize}
This step-by-step alternation ensures that the task difficulty (Scaffolder) and the solving capacity (Learner) evolve in exact lockstep, preventing catastrophic gradient divergence.

\textbf{Golden Experience Replay (Epoch-Level).}
After $N$ interleaved steps, we harvest the ``golden interactions'' from the successful trials. These consist of both the high-quality adaptive questions generated by the Scaffolder and the optimal reasoning traces produced by the Learner. We then format them as standard prompt-completion pairs $(x, y)$---where $x$ represents either the base context or the generated constraint and $y$ represents the desired output. 
These pairs are merged into a unified online replay buffer $\mathcal{D}_{\text{gold}}$. The unified model $\pi_\theta$ is then optimized via Supervised Fine-Tuning (SFT) cross-entropy loss over this mixed distribution:
\begin{equation}
    \mathcal{L}_{\text{SFT}}(\theta) = -\mathbb{E}_{(x, y) \sim \mathcal{D}_{\text{gold}}} \left[ \log \pi_\theta(y \mid x) \right]
\end{equation}

This combined architecture acts as a powerful self-stabilizing engine. While the step-level interleaved GRPO drives relentless exploration and boundary-pushing, the epoch-level Golden Experience Replay serves as a behavioral regularizer. It anchors the shared parameters in well-formed, highly logical trajectories, effectively mitigating the format collapse commonly observed in long-horizon reinforcement learning. Ultimately, consolidating the Learner's optimal trajectories inherently enhances the Scaffolder's precision in calibrating difficulty-aligned constraints, ensuring a continuous, self-sustaining upward spiral of capability.
\begin{table*}[t]
\centering
\small 
\setlength{\tabcolsep}{12.5 pt} 
\begin{tabular}{ll ccccc}
\hline
\multirow{2}{*}{\textbf{Model}} & \multirow{2}{*}{\textbf{Method}} & \multicolumn{2}{c}{\textbf{Mano Complex}} & & \multicolumn{2}{c}{\textbf{Complex OR}} \\
\cmidrule(lr){3-4} \cmidrule(lr){6-7}
 & & \textbf{ER (\%)} & \textbf{SA (\%)} & & \textbf{ER (\%)} & \textbf{SA (\%)} \\
\midrule
\multirow{8}{*}{Qwen2.5-7B-Instruct} 
 & Vanilla                 & 36.4 & 31.6 & & 16.7 & 10.5 \\
 & Vanilla + CoT           & 42.3 & 38.1 & & 25.6 & 19.1 \\
  \cmidrule{2-7}
 & ORLM                    & 57.1 & 50.6 & & 51.8 & 41.3 \\
 & LLMOPT                  & 62.7 & 54.1 & & 56.0 & 53.7 \\
 & StepORLM                & 65.8 & 62.4 & & 57.3 & 52.6
 \\
 & Absolute Zero           & 63.4 & 60.1 & & 56.8 & 54.1
 \\
& R-Zero                   & 64.8 & 61.7 & & 58.2 & 53.2
\\
 \cmidrule{2-7}
 & \textbf{STRETCH (Ours)} & \textbf{67.9} & \textbf{63.6} & & \textbf{58.7} & \textbf{55.0} \\
\midrule
\multirow{8}{*}{Llama-3.1-8B-Instruct}
 & Vanilla                 & 31.9 & 26.5 & & 15.3 & 11.4 \\
 & Vanilla + CoT           & 38.4 & 34.7 & & 30.6 & 25.3 \\
\cmidrule{2-7}
 & ORLM                    & 52.5 & 47.6 & & 46.1 & 42.3 \\
 & LLMOPT                  & 56.9 & 50.8 & & 52.2 & 47.4 \\
 & StepORLM                & 58.3 & 49.5 & & 53.6 & 45.8 \\
 & Absolute Zero           & 58.9 & 53.7 & & 53.3 & 48.4 \\
& R-Zero                   & 60.5 & 52.3 & & 55.3 & 46.8  \\
 \cmidrule{2-7}
 & \textbf{STRETCH (Ours)} & \textbf{61.2} & \textbf{54.8} & & \textbf{56.2} & \textbf{49.5} \\
\hline
\end{tabular}

\caption{Main experimental results on Operation Research (OR) tasks. We report Execution Rate (ER) and Solving Accuracy (SA) on both the Mano Complex and Complex OR, where the higher values of all reported metrics indicate superior performance.}
\label{tab:or_double_column_results}
\end{table*}
\section{Experiments}
\label{sec:experiments}
\subsection{Experimental Setup}
\label{subsec:exp_setup}
\noindent \textbf{Datasets and Evaluation Metrics.} 
To rigorously evaluate the generalizability and robustness of STRETCH, we select two constraint-solving domains: dynamic adversarial interactions and mathematical planning.
\begin{itemize}
    \item \textbf{Social Negotiation} (Soft Constraints): We evaluate on the \textit{CraigslistBargain} dataset \citep{he-etal-2018-decoupling}, a classic multi-turn buyer-seller negotiation scenario. In this domain, constraints (e.g., counterpart's bottom line and patience) are implicit and constantly shifting. Following standard protocols \citep{ahmad-etal-2023-ina,liu-etal-2025-dual}, we evaluate model performance using three metrics: \textit{Average Turn (AT)} (lower is better for efficiency), \textit{Success Rate (SR)} (rate of reaching a valid agreement), and \textit{Price Gap (PG)} (the fraction of the final selling price relative to the initial proposed price, indicating negotiation profitability).
    
    \item \textbf{Operation Research} (Hard Constraints): We utilize two hardcore mathematical programming benchmarks: \textit{Mano Complex} \citep{huang2024mamo} and \textit{Complex OR} \citep{xiao2024chainofexperts}. These tasks require the agent to parse real-world resource constraints into executable optimization code. The evaluation relies on objective programmatic metrics: \textit{Code Execution Rate  (ER)} (syntactic and runtime correctness) and \textit{Solving Accuracy (SA)} (whether the generated code yields the optimal objective value while strictly satisfying all mathematical constraints) \citep{JiangShu2025llmopt}.
\end{itemize}

\noindent \textbf{Baselines.} We benchmark our framework against both general-purpose Large Language Models and domain-specific methodologies. 
For the Negotiation task, we compare against (1) Prompt-based methods: Vanilla LLM, GDPZero \citep{yu-etal-2023-prompt}, and Pro-CoT \citep{deng-etal-2023-prompting}; and (2) Training-based methods: DPDP \citep{he-etal-2024-planning} and DMNA \citep{liu-etal-2025-dual}. 
For the Operation Research (OR) task, we evaluate against general reasoning baselines (Vanilla,  CoT) \citep{DBLP:journals/corr/abs-2201-11903} and specialized OR-LLM frameworks: ORLM \citep{huang2024orlm}, LLMOPT\citep{JiangShu2025llmopt}, and StepORLM \citep{zhou2026steporlm}. We further include controlled implementations of two closely related self-play approaches, Absolute Zero (AZR) \citep{zhao2025absolutezeroreinforcedselfplay} and R-Zero \citep{huang2025rzero}, using the same task contexts and evaluation metrics as STRETCH. In contrast to STRETCH's shared Scaffolder--Learner parameterization, epoch-level Golden Experience Replay, and explicit target difficulty, these methods do not instantiate this three-part design in our setting.

\noindent \textbf{Implementation Details.} 
Our unified Scaffolder-Learner framework is instantiated on two open-sourced base models: Qwen2.5-7B-Instruct \citep{qwen2025qwen25technicalreport} and Llama-3.1-8B-Instruct \citep{grattafiori2024llama3herdmodels}. 
During the micro-interleaved difficulty alignment and exploration phase, the Scaffolder generates $n=8$ adaptive questions, and the Learner samples $m=8$ trajectories per question. For the Negotiation tasks, the external LLM-as-a-Judge (GPT-4o-mini) \citep{openai2024gpt4technicalreport} evaluates deal outcomes to compute the rewards. For the OR tasks, an automated Python compiler along with GPT-4o-mini \citep{openai2024gpt4technicalreport} acts as the environment judge, returning binary rewards for execution and optimality. All experiments are conducted using LoRA \citep{hu2022lora} fine-tuning on two NVIDIA H100 GPUs.

For more experimental configuration details and related prompts, please refer to the Appendix~\ref{other_details} and Appendix~\ref{prompts}.

\subsection{Main Results}
\label{subsec:main_results}

The main experimental results across the Operation Research (OR) and Social Negotiation benchmarks are summarized in Table~\ref{tab:or_double_column_results} and Table~\ref{tab:negotiation_results}, respectively. Absolute Zero and R-Zero are reported in these main tables as direct self-play baselines. Their results are obtained from controlled implementations under the same task contexts and evaluation protocols as STRETCH; they should therefore be interpreted as in-domain comparisons rather than a universal ranking over the original papers' evaluation suites. Overall, STRETCH improves upon both general-purpose prompting methods and domain-specific training frameworks across two backbone models, demonstrating cross-domain generalization of the unified co-evolution paradigm.

\begin{table}[t]
\centering
 \small 
\setlength{\tabcolsep}{3.5pt} 
\resizebox{\columnwidth}{!}{
\begin{tabular}{ll ccc}
\hline 
\multirow{2}{1.6cm}{\textbf{Model}} & \multirow{2}{*}{\textbf{Method}} & \multicolumn{3}{c}{\textbf{CraigslistBargain}} \\
\cmidrule(lr){3-5}
& & \textbf{AT}  & \textbf{SR} & \textbf{PG} \\
\midrule 
\multirow{8}{1.6cm}{Qwen 2.5-7B\\-Instruct} 
 & Vanilla          & 10.18 & 0.33 & 0.83 \\
  \cmidrule{2-5}
 & GDPZero          & 9.36 & 0.54 & 0.87 \\
 & Pro-CoT          & 9.25 & 0.47 & 0.81 \\
 & DPDP             & 8.78 & 0.68 & 0.73 \\
 & DMNA             & 7.94 & 0.71 & 0.67 \\
 & Absolute Zero    & 7.88 & 0.67 & 0.75 \\
 & R-Zero           & 7.56 & 0.72 & 0.69 \\
 \cmidrule{2-5} 
 & \textbf{STRETCH} & \textbf{7.32} & \textbf{0.79} & \textbf{0.61} \\
\midrule 
\multirow{8}{1.6cm}{Llama-3.1-8B\\-Instruct}
 & Vanilla          &9.52 &0.41 &0.91 \\
  \cmidrule{2-5}    
 & GDPZero          &8.09 &0.57 &0.85 \\
 & Pro-CoT          &7.71 &0.69 &0.83 \\
 & DPDP             &\textbf{6.83} &0.65 &0.84  \\
 & DMNA             &7.36 &0.72 &0.76 \\
 & Absolute Zero    & 7.13 & 0.66 & 0.79 \\
 & R-Zero           & 7.46 & 0.73 & 0.84 \\
 \cmidrule{2-5}
 & \textbf{STRETCH} & 6.96 & \textbf{0.75} & \textbf{0.73} \\
\hline 
\end{tabular}
} 
\caption{Negotiation performance on CraigslistBargain dataset. Performance is evaluated via Average Turns (AT, $\downarrow$), Success Rate (SR, $\uparrow$), and Price Gap (PG, $\downarrow$).}
\label{tab:negotiation_results}
\end{table}

\textbf{Superior Performance on Hard Constraints (OR).} As shown in Table~\ref{tab:or_double_column_results}, pure reasoning baselines (Vanilla and Vanilla+CoT) struggle significantly with complex mathematical problems. While optimization-specific language models (e.g., ORLM, LLMOPT, and StepORLM) improve performance via static domain-specific fine-tuning, STRETCH further raises the upper bound. On the highly challenging Complex OR dataset, STRETCH built upon Qwen2.5-7B-Instruct achieves an Execution Rate (ER) of 58.7\% and a Solving Accuracy (SA) of 55.0\%, compared with 57.3\% ER and 52.6\% SA for StepORLM. It also exceeds the controlled AZR and R-Zero baselines on both Mano Complex and Complex OR. For example, on Mano Complex with Qwen2.5-7B-Instruct, STRETCH reaches 67.9\% ER and 63.6\% SA, compared with 63.4\% and 60.1\% for AZR and 64.8\% and 61.7\% for R-Zero. On the Llama-3.1-8B-Instruct backbone, STRETCH obtains the highest SA of 54.8\% on Mano Complex and 49.5\% on Complex OR. These results support the value of coupling adaptive difficulty with shared-role consolidation, rather than relying on self-play alone.

\textbf{Strategy Mastery in Soft Constraints (Negotiation).} 
Negotiation represents a highly adversarial environment where an agent must balance reaching an agreement (SR) with maximizing its own profit (represented by a lower Price Gap, PG) in as few turns as possible (AT). On the Qwen2.5-7B-Instruct backbone, STRETCH obtains the highest SR (0.79), the lowest PG (0.61), and the shortest AT (7.32). It improves over R-Zero by 0.07 SR and 0.08 lower PG, and over AZR by 0.06 SR and 0.07 lower PG. On Llama-3.1-8B-Instruct, STRETCH again achieves the best SR (0.75) and PG (0.73), while DPDP attains the shortest AT (6.83). Thus, the result is a success--profit advantage rather than a claim that STRETCH is best on every metric for every backbone. The lower PG indicates that the model does not merely compromise to reach agreement; its co-evolutionary training also improves bargaining outcomes.

In summary, 
the consistent superiority of STRETCH on both rigid mathematical planning (Table~\ref{tab:or_double_column_results}) and flexible strategic game play (Table~\ref{tab:negotiation_results}) empirically validates that our unified, dual-role parametric co-evolution mechanism can successfully break the capability stagnation inherent in static RL pipelines. For validating the generalization of STRETCH on simpler datasets, please refer to Appendix~\ref{sec:appendix_simple_or}.

\begin{figure*}[t]
    \centering
    \includegraphics[width=0.8\textwidth]{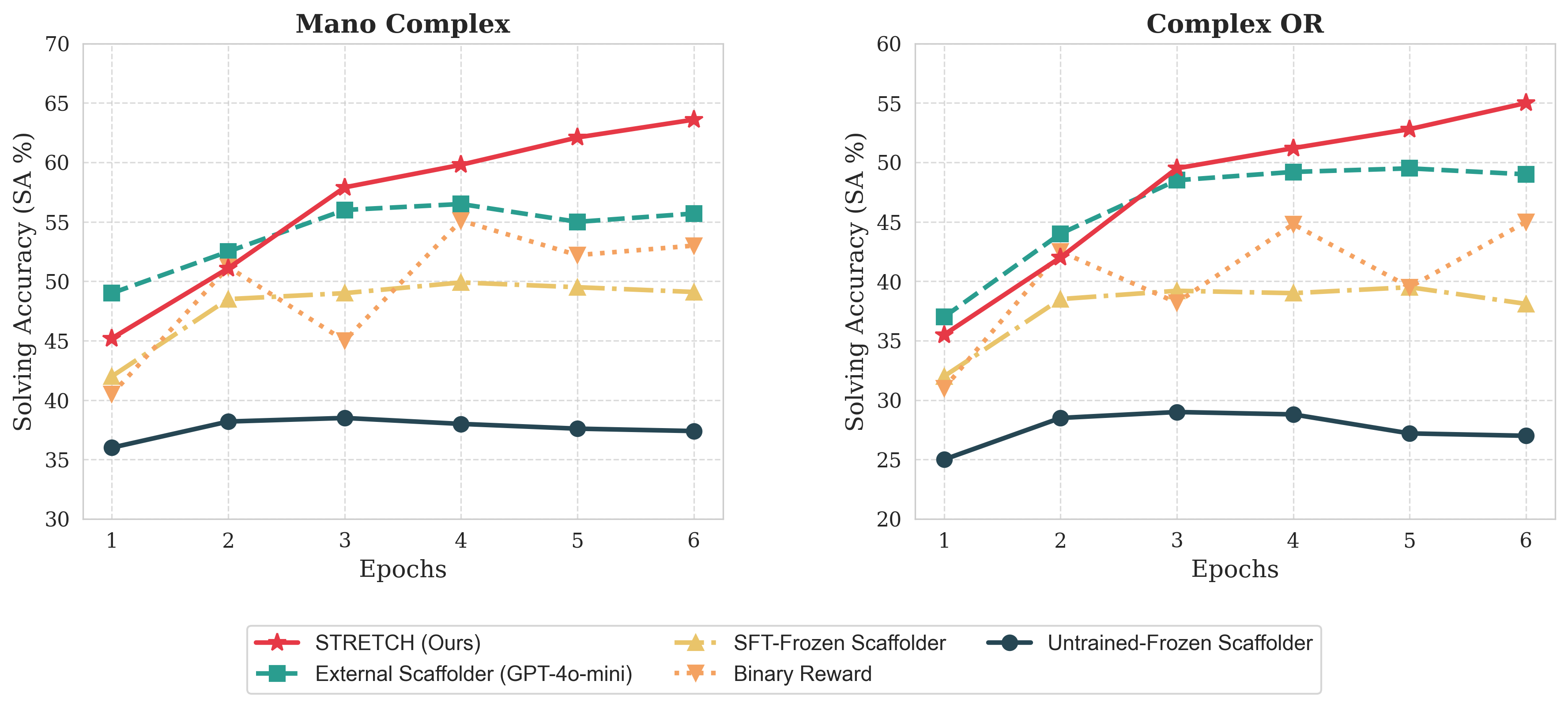}
    \caption{Ablation study on the necessity of difficulty alignment over a 6-epoch training on Learner by using the Qwen2.5-7B-Instruct backbone. We report Solving Accuracy (SA) on the Mano Complex (left) and Complex OR (right) datasets.}
    \label{fig:ablation_epochs}
\end{figure*}

\subsection{Ablation Studies: The Necessity of Difficulty Alignment}
\label{subsec:ablation}

To thoroughly investigate the necessity of dynamic difficulty alignment, we conduct an extensive ablation study over the co-evolution process. We compare STRETCH against following Scaffolder configurations using the Qwen2.5-7B-Instruct on the Mano Complex and Complex OR.
 
\textbf{Untrained-Frozen Scaffolder:} The Scaffolder is frozen at its initialized state without undergoing any updates, generating problems based solely on its base capabilities.
\textbf{SFT-Frozen Scaffolder:} The Scaffolder undergoes an initial SFT phase to learn basic constraint generation patterns but remains strictly frozen during the multi-epoch self-play.
\textbf{External Scaffolder (GPT-4o-mini):} We replace the internal co-evolving Scaffolder with an external model.
\textbf{Binary Reward:} The Scaffolder is dynamically updated, but $R_{diff}$ is replaced with a coarse binary signal. It receives a reward of 0 if the Learner answers all samples correctly or fails entirely, and 1 for any mixed accuracy.

The performance across 6 epochs (visualized in Figure~\ref{fig:ablation_epochs}) reveals critical insights into how task difficulty impacts solving capacity.

\begin{figure}[t]
    \centering
    \includegraphics[width=0.85\columnwidth]{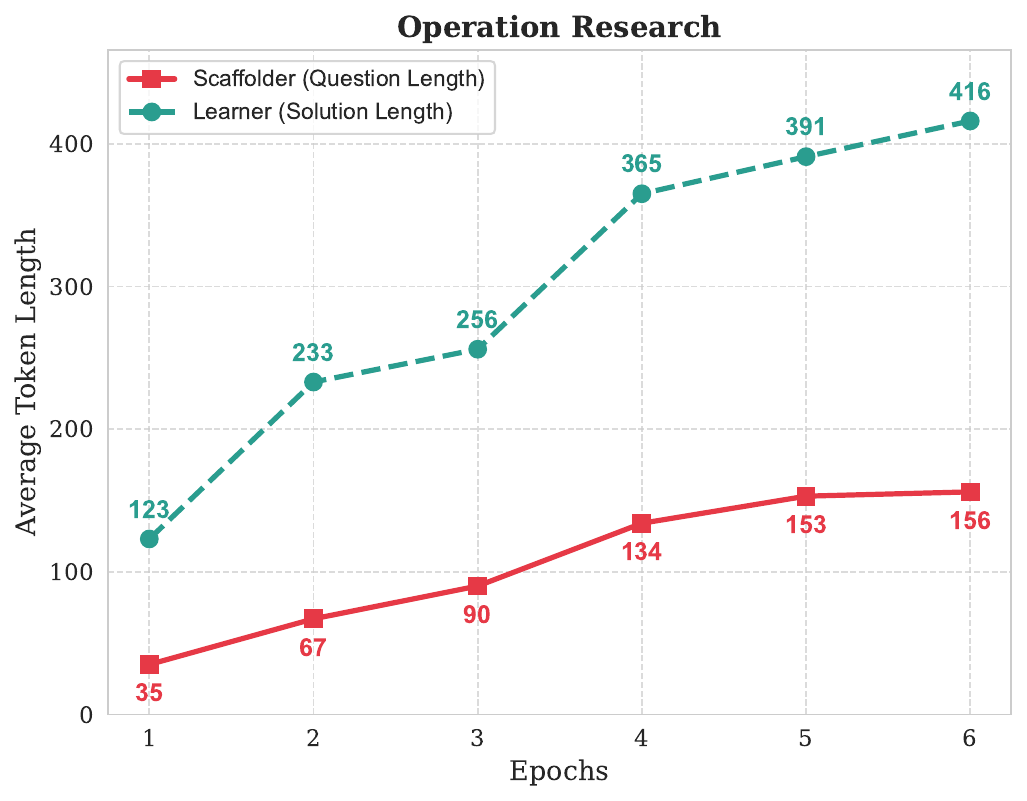}
    \caption{Quantitative co-evolution of question complexity (Scaffolder) and reasoning depth (Learner). The average token length of the generated questions and the successful reasoning trajectories exhibit a synchronized upward trend.}
    \label{fig:token_evolution}
\end{figure}

First, \textbf{static scaffolder inevitably leads to capability stagnation}. As observed in both benchmarks, the Untrained-Frozen variant serves as a lower bound. While the SFT-Frozen variant establishes a significantly higher initial baseline, its performance decisively flatlines after Epoch 2. Because the constraint difficulty remains fixed, the Learner rapidly exhausts the training value of the generated questions, collapsing into a Comfort Zone.

Second, \textbf{disjoint capabilities hinder long-term difficulty alignment}. Relying on the \textbf{External Scaffolder} yields strong early-epoch performance, as the proprietary model initially proposes highly challenging constraints. However, its trajectory noticeably plateaus midway. Because the external model cannot structurally internalize the Learner's specific cognitive boundaries and benefit from shared golden experience replay, it ultimately fails to calibrate its questions to the Learner's nuanced capacity shifts.

Finally, \textbf{the contrast between Binary Reward and STRETCH underscores the necessity of precise difficulty alignment}. Driven by a flat binary reward, the Binary Reward Scaffolder randomly oscillates between generating trivial and impossible constraints. This erratic behavior denies the Learner consistent gradient signals, resulting in severe performance variance and unpredictable drops. In stark contrast, STRETCH sustains a stable upward spiral. By utilizing the continuous reward $R_{diff}$, the Scaffolder is forcefully anchored to the exact edge of the Learner's capabilities. 

These ablation results empirically prove our core hypothesis: dynamic difficulty alignment within the Stretch Zone is the essential driver for continuous LLM evolution.

\begin{table*}[t]
\centering
\small
\setlength{\tabcolsep}{12pt} 
\begin{tabular}{llcccc}
\hline
\multirow{2}{*}{\textbf{Configuration}} &
\multirow{2}{*}{\textbf{Backbone}} &
\multicolumn{2}{c}{\textbf{Mano Complex}} &
\multicolumn{2}{c}{\textbf{CraigslistBargain}} \\
\cmidrule(lr){3-4} \cmidrule(lr){5-6}
& & \textbf{ER (\%)} & \textbf{SA (\%)} &
\textbf{SR} & \textbf{PG} \\
\midrule

\multirow{2}{*}{Separate LoRAs}
& Qwen2.5-7B-Instruct & 63.2 & 55.6 & 0.64 & 0.76 \\
& Llama-3.1-8B-Instruct & 57.8 & 52.6 & 0.67 & 0.82 \\
\cmidrule{2-6}
\multirow{2}{*}{\textbf{STRETCH (shared)}}
& Qwen2.5-7B-Instruct & \textbf{67.9} & \textbf{63.6} &
\textbf{0.79} & \textbf{0.61} \\
& Llama-3.1-8B-Instruct & \textbf{61.2} & \textbf{54.8} &
\textbf{0.75} & \textbf{0.73} \\

\midrule

No Golden Experience Replay
& Qwen2.5-7B-Instruct & 54.3 & 44.7 & 0.47 & 0.76 \\
\cmidrule{2-6}
Target $p=0.8$ (easy)
& Qwen2.5-7B-Instruct & 62.4 & 56.7 & 0.73 & 0.71 \\

Target $p=0.2$ (hard)
& Qwen2.5-7B-Instruct & 57.8 & 50.5 & 0.66 & 0.83 \\
\cmidrule{2-6}
\textbf{STRETCH ($p=0.5$)}
& Qwen2.5-7B-Instruct &
$\textbf{67.9}$ & $\textbf{63.6}$ &
$\textbf{0.79}$ & $\textbf{0.61}$ \\
\hline
\end{tabular}

\caption{Analysis of parameter sharing, Golden Experience Replay, and target difficulty on Mano Complex and CraigslistBargain.}
\label{tab:component_ablation}
\end{table*}
\subsection{Analysis of Co-Evolution Complexity}
\label{subsec:discussion_complexity}
To look into the Stretch Zone visually, we quantitatively analyze the structural evolution of both the Scaffolder's generated tasks and the Learner's successful solutions. We utilize the \textit{Average Token Length} as a proxy for cognitive complexity. For the Scaffolder, longer constraints indicate the intricate piecewise mathematical limits. For the Learner, longer successful trajectories reflect more extensive reasoning, profound logical planning, and complex code generation.

As illustrated in Figure~\ref{fig:token_evolution}, both roles exhibit a remarkable, synchronized upward trajectory across the 6-epoch evolution. In the Operation Research domain, as the Scaffolder's question descriptions grow from an average of 35 tokens to 156 tokens, the Learner's corresponding optimal code and reasoning traces expand from 123 tokens to over 416 tokens. This synchronized escalation provides definitive empirical evidence for parametric co-evolution. The Scaffolder is not simply generating impossible noise; rather, it systematically calibrates the difficulty. Concurrently, the Learner is not collapsing under harder constraints; instead, it matches the escalating difficulty by unlocking deeper reasoning capacities.

To provide a concrete, intuitive understanding of how this structural expansion translates into semantic difficulty, Table~\ref{tab:case_study} presents actual text segments generated by the STRETCH Scaffolder across different training epochs.

As observed in the examples, the Scaffolder does not merely pad the prompt with meaningless tokens to exploit the reward function. Instead, it systematically escalates the cognitive challenge, anchoring the constraints precisely within the Learner's Stretch Zone. For concrete qualitative examples illustrating how these token expansions translate into explicit mathematical logic and negotiation strategies, please refer to Appendix~\ref{sec:appendix_case_study}.

\subsection{Analysis of Shared Roles, Replay, and Target Difficulty}
\label{subsec:component_ablation}

Table~\ref{tab:component_ablation} examines three mechanisms that make the shared co-evolution process stable and productive. First, sharing one adapter between the Scaffolder and Learner is consistently stronger than maintaining separate role-specific LoRAs: it improves both OR execution/solving and negotiation success/profit on the two backbones. 

Second, Golden Experience Replay is essential for preserving useful behavior while the task distribution becomes harder. Without replay, performance drops sharply and the configuration exhibits format collapse by Epoch~4. Replaying high-quality trajectories therefore serves not only as a performance buffer, but also as a stabilizer for long-horizon self-play updates.

Finally, the boundary target $p=0.5$ is stronger than both an easier target ($p=0.8$) and a harder target ($p=0.2$). The easy setting supplies problems that are too often already solved, whereas the hard setting yields fewer useful successful trajectories. The balanced target achieves 67.9 ER and 63.6 SA on Mano Complex, together with 0.79 SR and 0.61 PG in negotiation. 
Together, these results support maintaining tasks near the current capability boundary.

\section{Conclusion}
In this paper, we introduced STRETCH, a unified framework for progressive self-improvement in verifiable, constraint-grounded LLM tasks. By conceptualizing the Stretch Zone and unifying the Scaffolder and Learner within a single parameter space, STRETCH replaces static training with a continuous, difficulty-aligned co-evolutionary loop. The controlled comparisons and component analyses show the benefits of shared role information, replay stabilization, and a boundary-level difficulty target on our OR and negotiation settings. These findings support STRETCH as a practical approach to self-improvement where reliable task generation and verification are available.
\section*{Limitations}
\textbf{Dependency on Reward Verifiability and Simulation.} The efficacy of the difficulty-alignment reward and the Learner's accuracy reward inherently relies on environmental feedback. In OR, invalid or infeasible constraints are deterministically rejected by the programmatic verifier. In negotiation, however, GPT-4o-mini serves as both a seller simulator and part of the reward/evaluation pipeline. This creates simulator dependence: improvements may partly reflect adaptation to that model's behaviors and preferences, and the service can change over time. We therefore limit our claims to settings with reliable task generation and verification, document prompts and deterministic conversation seeds in the released code, and design the pipeline so that GPT-4o-mini can be replaced by an alternative evaluator. Independent human or cross-simulator evaluation remains necessary to establish transfer beyond this setup.

\textbf{Generalization to Unconstrained Environments.} STRETCH is currently validated in domains where cognitive complexity and task constraints can be formally defined and objectively evaluated. Extending this parametric co-evolution to purely open-ended tasks—where objective difficulty gradients are inherently subjective, multi-dimensional, or lack a clear mathematical boundary—presents a broader challenge. Formulating a computable and universally applicable ``Stretch Zone'' for subjective, unconstrained domains is an open question that warrants further theoretical investigation.

\textbf{Scope of Claims.} Accordingly, our empirical claims are restricted to the evaluated, constraint-grounded settings; they do not establish general open-ended or autonomous LLM evolution.
\bibliography{custom}

\appendix
\section{Other Experiment Details}
\label{other_details} 

\subsection{Simulation and Reward Formulation for Negotiation}
Unlike static mathematical reasoning, negotiation is a dynamic, multi-turn adversarial game. To properly evaluate and optimize the Learner in the CraigslistBargain dataset, we deploy \texttt{GPT-4o-mini} as a rule-abiding simulator acting as the Seller, while our Learner takes on the role of the Buyer. 

To accurately estimate the value of each intermediate response generated by the Learner during the exploration phase, we adopt a rollout-based reward estimation mechanism, conceptually similar to CollabLLM\citep{wu2025collabllm}. Specifically, for each candidate reply generated by the Learner, we do not rely on a naive step-level heuristic. Instead, we simulate the remainder of the conversation to its conclusion by sampling multiple future trajectories (rollouts). The reward $R_{acc}$ assigned to the Learner's current reply is proportionally derived from the number of successful deals reached across these multiple simulations. If a reply consistently leads to a successful agreement within the target budget across the simulated futures, it receives a high reward; conversely, replies leading to negotiation breakdowns (walk-aways) are penalized. This Monte Carlo-style exploration ensures that the Learner optimizes for long-term strategic success rather than myopic, single-turn appeasement.

\subsection{Datasets and Training Configurations}
\textbf{Dataset Statistics.} For the Operations Research domain, we utilize the training splits of the Mano Complex and Complex OR datasets as the seed contexts for the Scaffolder. The Scaffolder generates constraints based on roughly 1,500 base linear and non-linear programming scenarios. For the Social Negotiation domain (CraigslistBargain), we utilize approximately 3,000 dialogue scenarios encompassing various item categories (e.g., housing, vehicles, electronics) to serve as the context conditions.

\textbf{Hyperparameters.} All models are fine-tuned using Low-Rank Adaptation (LoRA) to ensure memory efficiency. The LoRA rank is set to 16, with an alpha of 32 and a dropout rate of 0.05. We apply LoRA to all linear layers (q\_proj, k\_proj, v\_proj, o\_proj, gate\_proj, up\_proj, down\_proj). The training spans a total of 6 self-play epochs, which we empirically found sufficient to reach the capability plateau for both domains. We use the AdamW optimizer with a peak learning rate of 2e-5, accompanied by a cosine learning rate scheduler and a 3\% warmup ratio. During the Golden Experience Replay phase at the end of each epoch, the unified model is trained with a batch size of 16 for standard supervised fine-tuning (SFT).

\subsection{Detailed Description of Baselines}

\textbf{Negotiation Baselines:}
\begin{itemize}
    \item \textbf{Vanilla \& Pro-CoT:} Standard zero-shot prompting and Proactive Chain-of-Thought prompting, which injects planning steps before generating the dialogue response.
    \item \textbf{GDPZero:} A prompt-based framework utilizing Monte-Carlo Tree Search (MCTS) for goal-oriented dialogue policy planning without updating model weights.
    \item \textbf{DPDP \& DMNA:} State-of-the-art training-based frameworks. DPDP employs a dual-process planning mechanism for dialogue, while DMNA (Dual-Mind Negotiation Agent) fine-tunes models to balance strategic bottom lines and expressive conversational tactics.
\end{itemize}

\textbf{Operations Research (OR) Baselines:}
\begin{itemize}
    \item \textbf{Vanilla \& Vanilla+CoT:} Standard direct generation and step-by-step reasoning prompts for translating math word problems into optimization code.
    \item \textbf{ORLM:} A customized, training-based framework specifically designed for automated optimization modeling.
    \item \textbf{LLMOPT:} A recent method that learns to define and solve general optimization problems from scratch using structured formulations.
    \item \textbf{StepORLM:} The previous state-of-the-art that utilizes a self-evolving framework guided by generative process supervision, tailored for operations research language models.
\end{itemize}
\section{Qualitative Examples}
\label{sec:appendix_case_study}
In Section \ref{subsec:discussion_complexity}, we quantitatively demonstrated that the average token length of constraints generated by the Scaffolder increases significantly over the course of the 6-epoch self-play. 

In the Social Negotiation domain, the Scaffolder transitions from basic numerical boundaries (Epoch 1) to multi-turn adversarial tactics involving bundled conditions and persona shifts (Epoch 6). Similarly, in the Operations Research domain, it shifts from elementary linear inequalities to complex Mixed-Integer Linear Programming (MILP) logic—such as Big-M formulations for fixed charges—that inherently requires advanced mathematical modeling capabilities from the Learner. This qualitative evidence corroborates that STRETCH acts as an intelligent curriculum designer, fostering authentic strategy emergence.

\begin{table*}[hbp]
\centering
\small
\renewcommand{\arraystretch}{1.4} 
\begin{tabular}{p{0.08\textwidth} p{0.42\textwidth} p{0.42\textwidth}}
\toprule
\textbf{Epoch} & \textbf{Social Negotiation (Dynamic Soft Constraints)} & \textbf{Operations Research (Static Hard Constraints)} \\
\midrule

\textbf{Epoch 1}\newline \textit{(Comfort)} & 
\textbf{Generated Target:} ``Your target price is \$150. Do not accept anything above \$180. Be polite.'' \newline
\textit{Complexity:} Single numerical constraint with a basic persona. Easily solved by the base Learner. & 
\textbf{Generated Constraint:} ``Add a production capacity limit: The total number of units produced cannot exceed 500.'' \newline
\textit{Complexity:} Basic linear inequality ($\sum x_i \le 500$). Trivially mapped to executable code. \\
\midrule

\textbf{Epoch 3}\newline \textit{(Stretch)} & 
\textbf{Generated Target:} ``Your target price is \$130. You must persuade the seller to include free delivery. If they refuse, express disappointment and threaten to walk away.'' \newline
\textit{Complexity:} Multi-objective constraint requiring conditional dialogue acts and emotional shifts. & 
\textbf{Generated Constraint:} ``Add a piecewise cost constraint: The first 200 units cost \$5 each, and any additional units cost \$7 each due to overtime labor.'' \newline
\textit{Complexity:} Non-smooth piecewise objective requiring auxiliary variables and sequential logic parsing. \\
\midrule

\textbf{Epoch 6}\newline \textit{Boundary} & 
\textbf{Generated Target:} ``Your absolute limit is \$110. Employ a 'Good Cop/Bad Cop' persona. First, offer \$100 and heavily critique the item's condition. If rejected, reluctantly offer \$115 but strictly demand the original receipt and an extended 30-day warranty.'' \newline
\textit{Complexity:} Highly adversarial multi-turn strategic planning with strict preconditions and bundled tradeoffs. & 
\textbf{Generated Constraint:} ``Formulate a conditional integer constraint: If warehouse A is used (binary variable $y_A=1$), then at least 100 units must be stored there ($x_A \ge 100$), and the total logistics cost must incorporate a fixed setup fee of \$1000.'' \newline
\textit{Complexity:} Hardcore Mixed-Integer Linear Programming (MILP) requiring Big-M formulations. \\

\bottomrule
\end{tabular}
\caption{Qualitative evolution of the constraints generated by the STRETCH Scaffolder. As the training progresses, the Scaffolder autonomously learns to escalate the cognitive complexity, perfectly matching the Learner's growing capacity without relying on human-curated curriculum.}
\label{tab:case_study}
\end{table*}
\section{Generalization to Simpler OR Datasets}
\label{sec:appendix_simple_or}

To verify that our STRETCH framework does not suffer from catastrophic forgetting on fundamental problems while escalating cognitive difficulty, we further evaluate it on two relatively simpler Operations Research benchmarks: \textbf{NL4Opt} and \textbf{Mano Easy}.

\textbf{Dataset Descriptions:}
\begin{itemize}
    \item \textbf{NL4Opt:} \citep{pmlr-v220-ramamonjison23a} A widely used benchmark focused on extracting and formulating linear programming (LP) problems from natural language descriptions. It primarily consists of textbook-level, explicit LP problems with straightforward continuous variables and linear constraints, serving as a fundamental test for optimization modeling.
    \item \textbf{Mano Easy:} \citep{huang2024mamo} A subset of the MAMO benchmark containing elementary mathematical modeling scenarios. Unlike the Mano Complex dataset used in our main experiments, the Easy subset focuses on basic linear relationships without requiring complex conditional logic or Mixed-Integer Linear Programming (MILP) formulations.
\end{itemize}

\textbf{Results and Analysis:}

As shown in Table~\ref{tab:simple_or_results}, STRETCH achieves highly comparable, and in most cases slightly superior, performance on these elementary datasets relative to strong domain-specific baselines (e.g., StepORLM and LLMOPT). 

Crucially, these results address a common vulnerability in self-improvement and curriculum learning paradigms: \textit{catastrophic forgetting}. While the Scaffolder dynamically pushes the Learner into the upper bounds of its \textit{Stretch Zone} (escalating to complex, adversarial, and MILP formulations in later epochs as discussed in Section 4.4), the Learner completely retains its foundational reasoning skills. 

This retention is largely attributed to our epoch-level Golden Experience Replay mechanism. By consistently consolidating high-quality trajectories into the shared parameter space, STRETCH ensures that the model maintains near-ceiling performance on fundamental tasks while simultaneously expanding its upper capability limits on hardcore constraints.

\begin{table*}[hbp]
\centering
\small
\renewcommand{\arraystretch}{1.2}
\begin{tabular}{llcccc}
\toprule
\multirow{2}{*}{\textbf{Model}} & \multirow{2}{*}{\textbf{Method}} & \multicolumn{2}{c}{\textbf{NL4Opt}} & \multicolumn{2}{c}{\textbf{Mano Easy}} \\
\cmidrule(lr){3-4} \cmidrule(lr){5-6}
 & & \textbf{ER (\%)} & \textbf{SA (\%)} & \textbf{ER (\%)} & \textbf{SA (\%)} \\
\midrule
\multirow{6}{*}{\textbf{Qwen2.5-7B-Instruct}} 
 & Vanilla & 60.5 & 55.2 & 65.1 & 60.3 \\
 & Vanilla + CoT & 75.3 & 70.1 & 78.4 & 72.5 \\
 & ORLM & 88.5 & 85.1 & 89.2 & 86.4 \\
 & LLMOPT & 90.1 & 87.5 & 91.5 & 88.0 \\
 & StepORLM & \textbf{92.4} & 90.1 & 93.1 & 91.2 \\
 & \textbf{STRETCH (Ours)} & 91.8 & \textbf{90.6} & \textbf{93.6} & \textbf{91.5} \\
\midrule
\multirow{6}{*}{\textbf{Llama-3.1-8B-Instruct}}
 & Vanilla & 55.4 & 49.8 & 60.2 & 54.1 \\
 & Vanilla + CoT & 70.2 & 64.5 & 72.8 & 67.9 \\
 & ORLM & 84.3 & 80.6 & 85.4 & 81.2 \\
 & LLMOPT & 86.8 & 83.2 & 88.1 & \textbf{87.6} \\
 & StepORLM & 89.5 & \textbf{87.2} & 89.1 & 86.3 \\
 & \textbf{STRETCH (Ours)} & \textbf{90.1} & 86.4 & \textbf{90.3} & 88.4 \\
\bottomrule
\end{tabular}
\caption{Performance comparison on simpler OR datasets (NL4Opt and Mano Easy). STRETCH maintains near-ceiling performance, demonstrating that escalating challenge complexity during self-play does not lead to catastrophic forgetting of foundational reasoning skills.}
\label{tab:simple_or_results}
\end{table*}
\section{Related Prompts}
\label{prompts}

In this section, we detail the exact end-to-end prompt templates used to instantiate the Scaffolder and the Learner within our STRETCH framework across the two evaluated domains. The dual roles are activated within the singular parameter space purely via these instructions.

\subsection{Prompts for Social Negotiation}

For the CraigslistBargain dataset, the Scaffolder acts as a strategic director imposing specific bargaining targets, while the Learner acts as the actual buyer executing the dialogue.

\vspace{2mm}
\noindent\textbf{Scaffolder Prompt (Constraint Generation):}
\begin{quote}
\small
You are an expert negotiation strategist. Your goal is to generate a specific, challenging, but achievable negotiation constraint for a buyer. It must not be trivially easy (e.g., just asking for a 1\% discount) nor impossibly hard (e.g., asking for a 90\% discount). Based on the following item description and base price: \{Base\_Item\_Description\_and\_Price\}, generate a specific target and a behavioral constraint for the buyer. Output your constraint in a single concise paragraph.
\end{quote}

\vspace{2mm}
\noindent\textbf{Learner Prompt (Task Solving):}
\begin{quote}
\small
You are a buyer negotiating the price of an item with a seller. You must strictly adhere to your internal constraints and goals while maintaining a realistic conversational tone. The item you are negotiating for is: \{Base\_Item\_Description\_and\_Price\}. Your specific constraint is: \{Generated\_Constraint\_from\_Scaffolder\}. Here is the dialogue history so far: \{Dialogue\_History\}. Generate your next dialogue response to the seller to maximize your strategic advantage and achieve your constraint.
\end{quote}

\subsection{Prompts for Operations Research}

For the mathematical modeling tasks (Mano Complex and Complex OR), the Scaffolder acts as an environment designer adding realistic resource limits, while the Learner acts as the mathematical solver.

\vspace{2mm}
\noindent\textbf{Scaffolder Prompt (Constraint Generation):}
\begin{quote}
\small
You are an expert Operations Research (OR) professor. Your task is to increase the complexity of a basic linear programming problem by adding a realistic, mathematically sound constraint. This could include adding a piecewise cost function, a budget limit, or a conditional integer constraint (e.g., Big-M formulation). The added constraint must be logically consistent with the base problem. Given the following base math problem: \{Base\_Math\_Problem\}, generate ONE new constraint paragraph to be appended to the problem description. Do not solve the problem.
\end{quote}

\vspace{2mm}
\noindent\textbf{Learner Prompt (Task Solving):}
\begin{quote}
\small
You are an expert Operations Research solver. Your task is to read a complex mathematical word problem, formulate it accurately, and write executable Python code using the Gurobi or PuLP library to solve it. The base problem is: \{Base\_Math\_Problem\}. Please also consider the following additional constraints: \{Generated\_Constraint\_from\_Scaffolder\}. Provide a step-by-step mathematical reasoning (Chain-of-Thought) followed by the complete, executable Python code block.
\end{quote}

\subsection{Prompts for LLM-as-a-Judge (Reward Evaluation)}

To compute the empirical accuracy ($R_{acc}$) during the micro-interleaved GRPO updates, we utilize an external judge (\texttt{GPT-4o-mini}). Below is the end-to-end evaluation prompt for the Negotiation domain.

\vspace{2mm}
\noindent\textbf{Judge Prompt (Negotiation Evaluation):}
\begin{quote}
\small
You are an impartial negotiation referee. Read the following completed negotiation dialogue and the buyer's secret constraint. Dialogue: \{Completed\_Dialogue\_Trajectory\}. Buyer's Constraint: \{Generated\_Constraint\_from\_Scaffolder\}. Did the buyer successfully reach a deal with the seller that STRICTLY satisfies their secret constraint? Output only "1" for Yes, or "0" for No.
\end{quote}
\end{document}